\documentclass[conference]{IEEEtran}
\IEEEoverridecommandlockouts

\usepackage[T1]{fontenc}
\usepackage[utf8]{inputenc}

\usepackage{cite}
\usepackage{amsmath,amssymb,amsfonts}

\usepackage{graphicx}
\usepackage{array}
\usepackage{booktabs}
\usepackage{xcolor}
\usepackage{soul}
\usepackage{url}

\usepackage{algorithm}
\usepackage{algpseudocode}   

\usepackage[hidelinks]{hyperref}

\def\BibTeX{{\rm B\kern-.05em{\sc i\kern-.025em b}\kern-.08em
  T\kern-.1667em\lower.7ex\hbox{E}\kern-.125emX}}

\begin{document}

\title{An Exploratory Study on Crack Detection in Concrete through Human-Robot Collaboration}

\author{Junyeon Kim$^{1}$, Tianshu Ruan$^{1}$, Cesar Alan Contreras$^{1}$, and Manolis Chiou$^{2}$
\thanks{$^{1}$ Extreme Robotics Lab (ERL) and National Center for Nuclear Robotics (NCNR), University of Birmingham, UK.}%
\thanks{$^{2}$School of Electronic Engineering and Computer Science, Queen Mary University of London, UK (m.chiou@qmul.ac.uk).}%
}


\maketitle

\begin{abstract}
Structural inspection in nuclear facilities is vital for maintaining operational safety and integrity. Traditional methods of manual inspection pose significant challenges, including safety risks, high cognitive demands, and potential inaccuracies due to human limitations. Recent advancements in Artificial Intelligence (AI) and robotic technologies have opened new possibilities for safer, more efficient, and accurate inspection methodologies. Specifically, Human-Robot Collaboration (HRC), leveraging robotic platforms equipped with advanced detection algorithms, promises significant improvements in inspection outcomes and reductions in human workload. This study explores the effectiveness of AI-assisted visual crack detection integrated into a mobile Jackal robot platform. The experiment results indicate that HRC enhances inspection accuracy and reduces operator workload, resulting in potential superior performance outcomes compared to traditional manual methods.

\end{abstract}

\begin{IEEEkeywords}
crack detection, robotics, computer vision, nuclear infrastructure, human–robot interaction, human-robot collaboration
\end{IEEEkeywords}

\section{Introduction}
Nuclear energy plays a critical role in global power production, requiring stringent safety and inspection protocols to prevent catastrophic failures. With aging facilities, maintaining structural integrity through regular and precise inspections of nuclear facilities is crucial due to the potential consequences associated with structural degradation, such as radiation leaks and facility downtime. Manual inspection remains the traditional approach, relying heavily on human visual assessments supplemented by tools such as ultrasonic scanners or infrared thermography. Traditional manual inspection methods are effective for human assessment, but they have constraints and limitations, including safety hazards, human error, substantial cognitive loads, and financial burdens \cite{b10}.

Emerging technologies, specifically Artificial Intelligence (AI) and robotics, provide promising alternatives by mitigating human risks, reducing cognitive strain, and enhancing accuracy and efficiency in structural inspections \cite{b10}. Advances in computer vision, particularly deep learning models such as Convolutional Neural Networks (CNNs) \cite{b10} and real-time object detectors such as You Only Look Once (YOLO) \cite{b1,b2}, have improved the detection and assessment of structural anomalies such as cracks. Integrating these AI models into robotic systems further enhances the potential for autonomous or semi-autonomous inspections, improving safety and efficiency in hazardous environments \cite{b10}.

This paper focuses on the integration of HRC within the scope of nuclear inspection, specifically examining how AI-assisted robotic systems influence human-robot team performance and workload during crack inspection tasks. By employing a mobile robotic platform equipped with the YOLOv8 detection model, we conduct empirical evaluations to determine whether this collaborative approach reduces human workload and can enhance inspection performance. This study thus aims to provide some evidence supporting the deployment of AI-driven robotic systems within nuclear inspection operations, emphasizing larger operational efficiency, increased safety, and higher accuracy.

In Table \ref{tab:comparison}, we present a comparison of crack detection methods, showing the advantages of employing mobile robotic systems integrated with AI-based detection models in nuclear inspection contexts.

\begin{table}[htbp]
  \caption{Comparative analysis of crack‑detection methods}
  \label{tab:comparison}
  \centering
  \setlength{\tabcolsep}{3pt} 
  \begin{tabular}{|m{0.18\linewidth}|m{0.28\linewidth}|m{0.38\linewidth}|}
    \hline
    \textbf{Method} & \textbf{Advantages} & \textbf{Disadvantages} \\ \hline
    \raggedright Manual inspection &
      High accuracy; direct human assessment &
      Slow; costly; hazardous \\ \hline
    \raggedright Fixed‑camera monitoring &
      Continuous monitoring capability &
      Limited coverage; dependent on human re-inspection \\ \hline
    \raggedright \textbf{Robotic inspection}\\
    (drones or mobile robots) &
      Real‑time detection; flexibility; adaptable to large outdoor and confined indoor structures &
      Requires robust localisation; stable communication; extensive, domain‑specific training datasets \\ \hline
  \end{tabular}
\end{table}
 
\section{Related Work}

\subsection{Image Processing and Detection Algorithms}
Visual crack detection traditionally involved manual or semi-automatic image processing techniques. Manual visual detection requires expert knowledge and experience in the field while semi-automatic image processing techniques rely on methods including edge detection, thresholding, and morphological filtering \cite{b10}. Recent advancements have introduced more accurate and efficient methods based on machine learning, notably CNNs \cite{b9}. CNNs offer high accuracy in detecting structural defects, though often at a computational cost limiting real-time application \cite{b9}.  

Among contemporary methods, the YOLO model stands out due to its real-time capabilities and single-pass inference approach, enabling easy and practical integrations with robotic systems requiring rapid and accurate visual feedback \cite{b1,b2}. YOLO models, particularly YOLOv8, achieve an optimal balance of detection speed and accuracy, making them suitable for real-time inspection tasks \cite{b2}.

\subsection{Robotic Systems in Nuclear Inspection}

Robotic platforms used in nuclear inspection scenarios generally fall into three broad categories: \textit{aerial} (drones), \textit{quadruped}, and \textit{wheeled robots}.  

Aerial drones excel at rapid coverage of large or vertical structures; however, their limited manoeuvrability and susceptibility to loss of positional accuracy make confined nuclear spaces challenging \cite{b11}.  
Quadruped robots, exemplified by Boston Dynamics’ Spot, provide superior stability on uneven terrain and can negotiate tight passages, making them attractive for hazardous indoor inspection tasks \cite{b12}.  
Wheeled platforms such as the Clearpath Jackal offer a reliable compromise of payload capacity, ease of localisation, and autonomous navigation in cluttered corridors \cite{b10}.  
Given these characteristics, our study employs the Jackal platform integrated with YOLOv8 to examine practical efficacy in AI‑assisted crack detection.

\subsection{AI-Based Robotic Systems Considerations}
Integrating AI-based object detectors such as YOLOv8 into inspection robots introduces several practical considerations beyond model accuracy. Compared with hand-engineered image-processing pipelines, learned detectors can be deployed rapidly and adapted to new data \cite{b9}. However, on-board execution requires careful optimisation to meet real-time constraints, including choices of input resolution, quantisation, and batching tailored to the available compute. The YOLOv8 family provides multiple model sizes (e.g., n/s/m/l/x) to trade accuracy for speed as needed \cite{b2}.

Detection quality depends strongly on data acquisition: camera calibration, exposure and lighting control, standoff distance, and viewpoint planning. Consistent capture conditions reduce both false alarms and misses \cite{b9}. Software integration must also synchronise sensor streams with the control stack; robust time-stamping and buffering between the vision process and robot control help maintain low-latency overlays and stable teleoperation \cite{b10}.

In this work, we define \emph{real-time} feedback to mean an operator-visible overlay rate of at least 10\,fps with end-to-end latency (camera capture to on-screen overlay) below 200\,ms, which is sufficient for responsive teleoperation and visual search.

\section{Experimental Setup and Procedures}
\begin{figure}[htbp]
\centering
\includegraphics[width=0.97\linewidth]{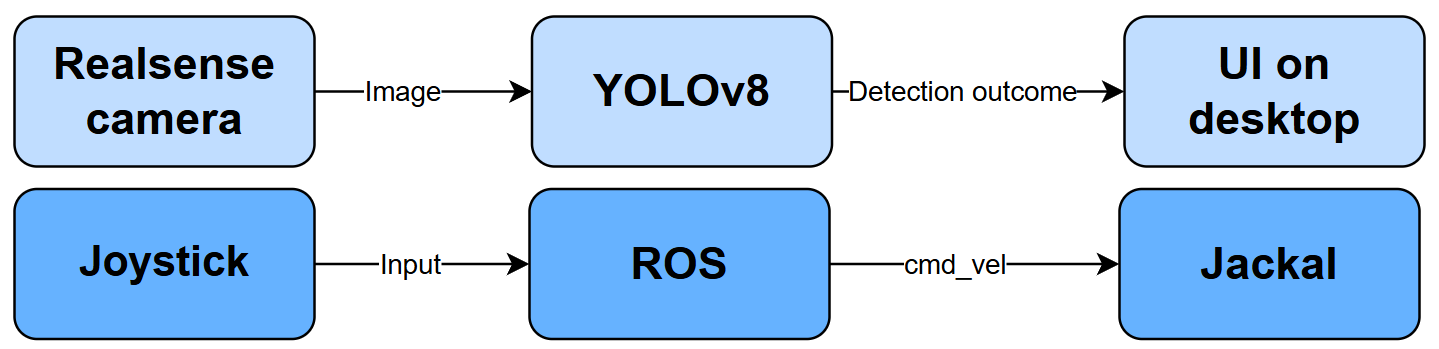}
\caption{System structure with two parallel threads: (top) inspection path
(RealSense $\rightarrow$ YOLOv8 $\rightarrow$ operator interface) and (bottom) teleoperation path
(joystick $\rightarrow$ ROS $\rightarrow$ \texttt{/cmd\_vel} $\rightarrow$ Jackal). 
YOLOv8 runs outside ROS; ROS does not feed the detector.}
\label{fig:architecture}
\end{figure}

The comparative experiment investigates whether HRC in crack detection tasks can enhance detection accuracy, reduce cognitive load, and improve overall efficiency in nuclear infrastructure inspections. Prior to the user study, YOLOv8 was trained on 500 images for 150 epochs; the resulting validation score was 81.6\% (Ultralytics-reported mAP@0.5). Training was performed on Ubuntu 22.04 with an NVIDIA GTX 1080 GPU and an Intel Core i7 CPU. To minimise end-to-end latency, the AI-assisted crack-detection loop ran outside ROS; the system therefore used two parallel threads—one for detection and one for robot navigation (Fig.~\ref{fig:architecture}).
 Before the experiment, YOLOv8 training was performed. For this 500 images were collected, and the training went on for 150 epochs, reaching an accuracy of 81.6\%, on an Ubuntu 22.04 desktop with an NVIDIA 1080 GTX GPU, i7 processor.

To minimise end-to-end latency, the AI-assisted crack-detection loop ran outside ROS; the system therefore used two parallel threads—one for detection and one for robot navigation (Fig.~\ref{fig:architecture}).

\paragraph*{Latency criterion and verification}
We instrumented timestamps at (i) camera-frame arrival, (ii) post–YOLO inference, and (iii) GUI overlay. 
We targeted \emph{real-time} feedback defined as an overlay update rate of $\geq\,10\,\mathrm{fps}$ and end-to-end latency $<\,200\,\mathrm{ms}$ (camera $\rightarrow$ overlay). 
During pilot runs, the inference queue depth remained at or below one frame, and no backlog was observed in the GUI overlay. Subsequently, we introduce an arena setup and procedures as follows:

\subsection{Arena Setup}
\begin{figure}[htbp]
\centering
\includegraphics[width=0.99\linewidth]{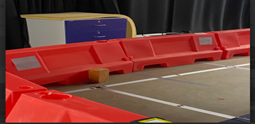}
\caption{Red barriers delineate the test area in which the Jackal robot operated. Images on the barriers mock up the cracked and uncracked concrete.}
\label{fig:setup}
\end{figure}

\begin{figure}[htbp]
\centering
\includegraphics[width=0.99\linewidth]{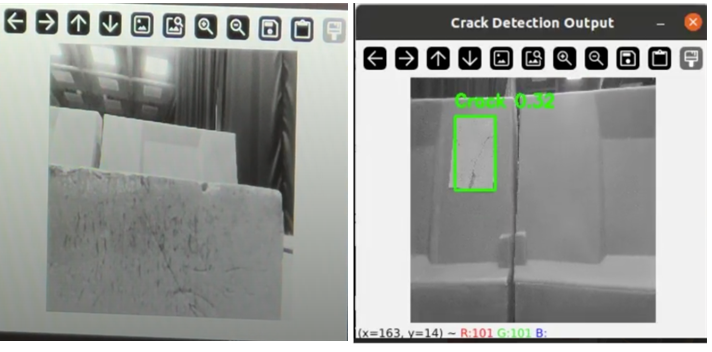}
 \caption{Operator’s video feed under the two experimental conditions. %
  \textbf{Left}: UI of trial with manual inspection, showing an un‑augmented camera view. %
  \textbf{Right}: UI of trial with AI‑assisted inspection, where the YOLOv8 model overlays a green bounding box and confidence score around the detected crack.}
\label{fig:video}
\end{figure}

The closed rectangular arena consists of four walls of red barriers. In the arena, printed images simulating cracked and un-cracked concrete surfaces were placed on barriers, forming an enclosed inspection area (Fig. \ref{fig:setup}). In each trial, the places of the image that refer to a cracked or uncracked concrete surface are swapped. Participants observed the scenario through the live video feed from the robot's camera, as shown in Fig. \ref{fig:video}.

\subsection{Implementation}
Six participants each completed two trials, one manual detection and one AI-assisted detection, each lasting three minutes. No further repetitions were performed; the single run for each condition was deemed sufficient because the test area was small and the crack distribution was fixed, allowing participants to cover the entire inspection area in the allotted time.

Before the experiment, each participant was trained to visually identify cracks on concrete surfaces using sample images that were not included in the actual experiment. Following this training, participants underwent a brief two-minute familiarisation session to practice robot driving and camera-view navigation. This helped to guarantee each participant had a similar understanding of the given tasks. Data from this session was not recorded. 

After the familiarisation session, participants completed two timed experimental trials. In each trial, participants navigated the robot from a starting point to an endpoint while simultaneously identifying and counting cracks. In each trial, the participant was given camera views on a UI on the screen with different information: \textit{Trial A} presented a raw, unaugmented video feed, while \textit{Trial B} provided AI-assisted notifications highlighting detected cracks running the trained YOLOv8 model, (see Fig. \ref{fig:video}). 

In \textit{Trial A} (manual inspection), the participant teleoperated the Jackal for three minutes and counted visible cracks while viewing a raw video feed on the monitoring screen. After that, the participant reported the total number of cracks observed, and the experimenter recorded this number.

In \textit{Trial B} (AI-assisted inspection), the same three-minute procedure was repeated, but participants viewed video streams augmented with green bounding-box overlays generated by the YOLOv8 model. Participants were instructed to confirm true positives, note any misclassifications, and provide a final crack count at the end of the trial. To mitigate potential learning effects from the previous trial, the positions of cracked and uncracked images were rearranged before each trial.

Immediately after each trial, the participant completed a NASA Task Load Index (NASA‑TLX) \cite{b3} questionnaire, rating mental demand, physical demand, temporal demand, effort, frustration, and perceived performance. The trial order was counterbalanced: three of the six participants performed the AI‑assisted trial first, while the rest began with the manual trial. All robot data, video frames, YOLO inference outputs, crack counts, and NASA TLX scores were recorded for subsequent analysis.

\section{Results and Analysis}
All six participants’ objective \textit{detection accuracy} (the percentage of cracks correctly identified and counted) was recorded, while subjective \textit{workload} was evaluated with the NASA‑TLX \cite{b3}, which captures mental, physical, and temporal demands, perceived performance, effort, and frustration. Improvement is therefore defined as higher detection accuracy together with lower NASA‑TLX scores. Given the small sample size ($n{=}6$) and the preliminary nature of this study, we rely on descriptive statistics and visual inspection of trends rather than formal inferential tests. Hence, we present the results from two points of view: (i) objective detection accuracy and (ii) subjective workload.

\subsection{Detection Accuracy}

\begin{figure}[t]
\centering
\includegraphics[width=.95\columnwidth]{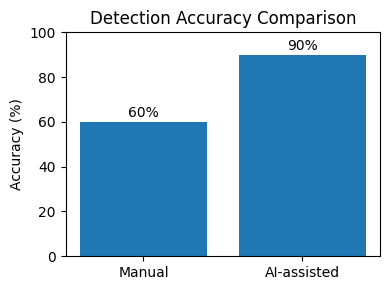}
\caption{Crack‑detection accuracy for each participant under manual and AI‑assisted conditions.}
\label{fig:accuracy}
\end{figure}

Fig. \ref{fig:accuracy} shows the distribution of crack‑detection accuracy for the manual and AI‑assisted conditions. All six participants displayed higher accuracy when guided by detection visual cues, with mean performance increasing from 60\% to 90\%. Although no statistical test is reported here, the non‑overlapping ranges suggest a performance gain worth validating in future studies with larger samples.

\subsection{NASA TLX Workload Profiles}

Table \ref{tab:nasatlx} compares the mean of NASA TLX scores (0–10 scale) in six workload dimensions. Specifically, from the scope of \textit{Mental demand}, it shows the largest decline from 7.8 (manual) to 3.2 (AI-assisted). The \textit{physical} demand remains steady with 6.0 in both trials, reflecting identical robot control with a joystick in both trials. \textit{Perceived Performance} is observed to increase from 6.0 (manual) to 9.0 (AI-assisted). \textit{Temporal} demand and \textit{Effort} drop from 7.5 (manual) to 4.1 (AI-assisted) and from 7.2 (manual) to 4.5 (AI-assisted) separately. \textit{Frustration} ratings remained moderate and almost unchanged between conditions (manual: 5.0; AI‑assisted: 4.8). Post‑trial debriefs revealed two opposing influences: participants appreciated the bounding box cues that sped up detection, yet occasional false positives and brief display latency distracted them from driving tasks, offsetting potential gains in satisfaction. Previous HRI studies report similar tension, where assistance can reduce workload, but mistrust triggered by false alarms sustains frustration \cite{b9}. Addressing model precision and offering transparent confidence indicators, therefore, remain key design targets for future iterations.

Beyond workload reductions, AI assistance consistently improved crack‑detection accuracy for every participant. The mean accuracy increased from approximately 60\% to 90\%, a gain of 30 percentage points (about 50\% relative).
 This systematic gain, alongside the workload findings, stresses the practical value of HRC in inspection scenarios.




\begin{table}[t]
\caption{NASA TLX Mean Scores ($n{=}6$)}
\label{tab:nasatlx}
\centering
\renewcommand{\arraystretch}{1.25} 
\begin{tabular}{lcc}
\toprule
\textbf{Metric} & \textbf{Manual} & \textbf{AI‑Assisted} \\
\midrule
Mental Demand         & 7.8 & 3.2 \\
Physical Demand       & 6.0 & 6.0 \\
Temporal Demand       & 7.5 & 4.1 \\
Perceived Performance & 6.0 & 9.0 \\
Effort                & 7.2 & 4.5 \\
Frustration           & 5.0 & 4.8 \\
\bottomrule
\end{tabular}
\end{table}


\section{Discussion}
The experimental evidence indicates that a carefully designed collaboration between a human operator and an AI-enabled robotic platform improves inspection accuracy while easing the mental burden associated with continuous visual search. Two factors appear central to this improvement.

\textbf{First}, the AI provides a real-time stream of bounding boxes and confidence scores that enlarges the operator’s perceptual field, allowing rapid confirmation or rejection of potential defects without exhaustive manual scanning. By sharing a single perceptual workspace, the robot’s fast pattern recognition capabilities complement the operator’s contextual judgement, creating cooperation rather than competition. \textbf{Second}, routine inspection tasks are offloaded to the robot, freeing the operator’s cognitive resources for higher-level activities such as path planning, safety checks, and interpretation of ambiguous cases. The observed reduction in NASA‑TLX mental demand and effort scores indicates that a portion of the visual and memory workload may have been offloaded to the AI assistance. Although this trend is encouraging for applications in controlled environments, the present pilot data (six participants, short trials) are insufficient to draw operational safety conclusions; larger samples and longer‑duration field studies will be required to confirm the magnitude and persistence of this effect.

Despite these advantages, frustration ratings remained essentially unchanged between manual and AI‑assisted conditions. Post‑trial debriefs revealed three recurring irritants: (i) false positives that required additional verification, (ii) brief display latency, and (iii) the absence of transparent confidence cues. Prior HRI research shows that adaptive feedback and confidence visualisation can maintain calibrated trust in human‑robot teams \cite{b4}.  Incorporating such features could therefore reduce residual frustration and foster deeper operator reliance on the system.  

\section{Conclusion and future work}
This exploratory study shows that integrating a YOLOv8 crack‑detection model with a Jackal mobile robot can raise detection accuracy from 60\% to 90\% and halve self‑reported mental workload in a six‑participant pilot.  The results demonstrate the promise of HRC for safer, faster inspections in nuclear environments, where restricted access and radiation constraints limit manual work. 


Based on the insights and findings from the experiment, three strands appear most urgent for turning this pilot into an operational tool:  
(i) \textit{Data expansion:} collecting domain‑specific images under varied lighting and concrete types to sharpen model precision and reduce false alarms;  
(ii) \textit{Interface refinement:} adding confidence bars, region‑of‑interest heat maps, and adaptive alert thresholds to foster calibrated trust;  
(iii) \textit{Longitudinal field trials:} multi‑week deployments that track trust dynamics, skill retention, and fatigue in real‑world nuclear facilities.

\end{document}